\documentclass[conference]{IEEEtran}
\IEEEoverridecommandlockouts
\usepackage{cite}
\usepackage{amsmath,amssymb,amsfonts}
\usepackage{algorithmic}
\usepackage{graphicx}
\usepackage{textcomp}
\usepackage{xcolor}

\makeatletter
\newcommand{\removelatexerror}{\let\@latex@error\@gobble}
\makeatother
\usepackage{makecell}
\usepackage{bbding}
\usepackage{booktabs}
\usepackage{graphicx}
\usepackage{tabularx}
\usepackage{subcaption}
\usepackage{setspace}
\usepackage{url}
\usepackage{shuffle}

\usepackage{arydshln}

\def\BibTeX{{\rm B\kern-.05em{\sc i\kern-.025em b}\kern-.08em
    T\kern-.1667em\lower.7ex\hbox{E}\kern-.125emX}}
\usepackage{xspace}
\newcommand{\method}{TCM-FTP\xspace}
\newcommand{\dataset}{\textit{DigestDS}\xspace}

\usepackage{xspace}
\usepackage{xcolor}

\definecolor{SHINSYU}{HTML}{AB3B3A}

\usepackage[hidelinks]{hyperref}

\newcommand{\VspaceBefore}{\vspace{-1.5mm}}
\newcommand{\VspaceAfter}{\vspace{-1.5mm}}

\usepackage{multirow}
\usepackage{float}
\usepackage{enumitem}

\begin{document}

\title{\method: Fine-Tuning Large Language Models for Herbal Prescription Prediction
% \thanks{We gratefully acknowledge the support of NVIDIA Corporation with the GPU computational resources used for this research. }
% \thanks{This work was supported by the Natural Science Foundation of Beijing (No. L232033), the National Natural Science Foundation of China (No. 82374302), and the Science and Technology Innovation Project of China Academy of Traditional Chinese Medicine (No. CI12021A00513).}
% \thanks{Xingzhi Zhou and Xin Dong have contributed equally to this work.}
}

\author{
\IEEEauthorblockN{
 Xingzhi Zhou\IEEEauthorrefmark{1}\IEEEauthorrefmark{3}\IEEEauthorrefmark{7}\thanks{\IEEEauthorrefmark{7}Xingzhi Zhou and Xin Dong have contributed equally to this work.}\thanks{\IEEEauthorrefmark{3}This work was done while Xingzhi Zhou was an intern at NVIDIA.}, Xin Dong\IEEEauthorrefmark{2}\IEEEauthorrefmark{7}, Chunhao Li\IEEEauthorrefmark{1}, Yuning Bai\IEEEauthorrefmark{4}, Yulong Xu\IEEEauthorrefmark{5}, Ka Chun Cheung\IEEEauthorrefmark{3}, Simon See\IEEEauthorrefmark{6} \\ Xinpeng Song\IEEEauthorrefmark{2},  \Envelope Runshun Zhang\IEEEauthorrefmark{3}, \Envelope Xuezhong Zhou\IEEEauthorrefmark{2}, and \Envelope Nevin L. Zhang\IEEEauthorrefmark{1}\thanks{\Envelope  Corresponding authors.}
}
\IEEEauthorblockA{\IEEEauthorrefmark{1}Department of Computer Science and Technology, \\ The Hong Kong University of Science and Technology, Hong Kong, China\\\{xzhoubl, chunhao.li\}@connect.ust.hk,  lzhang@cse.ust.hk}
\IEEEauthorblockA{\IEEEauthorrefmark{2}Institute of Medical Intelligence, Department of Artificial Intelligence, Beijing Key Lab of Traffic Data Analysis and Mining, \\ School of Computer Science \& Technology, Beijing Jiaotong University, Beijing, 100044, China\\ \{x\underline{~}dong, xpsong, xzzhou\}@bjtu.edu.cn}
\IEEEauthorblockA{\IEEEauthorrefmark{4}Guang’anmen Hospital, China Academy of Chinese Medical Sciences, Beijing, China\\ byn-1973@163.com, runshunzhang@139.com}
\IEEEauthorblockA{\IEEEauthorrefmark{3}NVAITC, NVIDIA, Hong Kong, China\\ chcheung@nvidia.com}
\IEEEauthorblockA{\IEEEauthorrefmark{6}NVAITC, NVIDIA, Singapore \\  ssee@nvidia.com}
\IEEEauthorblockA{\IEEEauthorrefmark{5}School of Information Technology, Henan University of Chinese Medicine, Henan, China\\ flyxyl@126.com}

}
\maketitle

\begin{abstract}
% abstract 1.3 revised from Prof. Nevin.

Traditional Chinese medicine (TCM) has relied on specific combinations of herbs in prescriptions to treat various symptoms and signs for thousands of years. Predicting TCM prescriptions poses a fascinating technical challenge with significant practical implications. However, this task faces limitations due to the scarcity of high-quality clinical datasets and the complex relationship between symptoms and herbs. To address these issues, we introduce \textit{DigestDS}, a novel dataset comprising practical medical records from experienced experts in digestive system diseases. We also propose a method, TCM-FTP (TCM Fine-Tuning Pre-trained), to leverage pre-trained large language models (LLMs) via supervised fine-tuning on \textit{DigestDS}. Additionally, we enhance computational efficiency using a low-rank adaptation technique. Moreover, TCM-FTP incorporates data augmentation by permuting herbs within prescriptions, exploiting their order-agnostic nature. Impressively, TCM-FTP achieves an F1-score of 0.8031, significantly outperforming previous methods. Furthermore, it demonstrates remarkable accuracy in dosage prediction, achieving a normalized mean square error of 0.0604. In contrast, LLMs without fine-tuning exhibit poor performance. Although LLMs have demonstrated wide-ranging capabilities, our work underscores the necessity of fine-tuning for TCM prescription prediction and presents an effective way to accomplish this.

\end{abstract}


\section{INTRODUCTION}
% \# New building intro.

% \begin{itemize}
% \item TCM is important with many research and application. Application A, B, C. One of main approach in TCM is prescription.

% \item TCM prescription defined as: A->B as an important way to solve disease cure.  Prescription require long-time training in practice, leading to a challenge for scarse doctor for cure resources and new doctor training. 

% \item TCM prescription prediction is significant. Existing method: A,B,C. However, existing methods in prescription prediction of TCM do not fully model the complex relationships between TCM prescription and symptoms. 

% \item To address these challenges above, we propose XXX method. 
% \end{itemize}
% \textbf{TODO: polish intro.}

% Introduction: v1.1 

% Traditional Chinese Medicine (TCM) has been an indispensable part of healthcare for the Chinese population for thousands of years. TCM employs many practices, including herbal medicine, acupuncture, cupping therapy, and tuina massage \cite{cheungTCMMadeChina2011}. Herbal medicine is the primary treatment modality of TCM. It has been shown to effectively treat the novel coronavirus (COVID-19), resulting in improved cure rates and reduced mortality \cite{liu2020efficacy, niCombatingCOVID19Integrated2020}.
Traditional Chinese Medicine (TCM), integral to Chinese healthcare for millennia, encompasses practices such as herbal medicine, acupuncture, cupping therapy, and tuina massage. As the primary modality of TCM, herbal medicine has demonstrated effectiveness in treating COVID-19, leading to improved cure rates and reduced mortality \cite{liu2020efficacy}.
% Herbal prescriptions require doctors to assess patient symptoms using the four diagnostic methods: observation (wang), listening and smelling (wen), questioning (wen), and pulse-taking (qie), guided by the principle of li-fa-fang-yao \cite{huang2012exploring}. Training a doctor involves extensive experience and continuous practical feedback, resulting in a prolonged training period. This contributes to a shortage of TCM doctors and limited TCM resources. Developing an effective prescription predictor is an approach to alleviate challenges such as doctor shortages and the need for prolonged training periods. 
Herbal prescriptions in TCM require doctors to assess patient symptoms using four diagnostic methods—observation (wang), listening and smelling (wen), questioning (wen), and pulse-taking (qie)—guided by the principle of li-fa-fang-yao \cite{huang2012exploring}. The extensive experience and continuous practical feedback required for training result in prolonged training periods, contributing to a shortage of practitioners and limited TCM resources. Developing an effective prescription predictor could alleviate challenges such as doctor shortages and lengthy training requirements.

 % Computational systems for prescription prediction in TCM involve designing a computational system capable of predicting the appropriate prescription based on given symptoms. This system models the complex relationships between symptoms and herbs \cite{yao2018topic}. Language generation techniques show particular promise in TCM prescription prediction, treating prescription generation as a machine translation problem and solving it using a sequence-to-sequence (seq2seq) model  \cite{sutskeverSequenceSequenceLearning2014}. After numerous attempts by various researchers \cite{liuAttentiveherbNovelMethod2019,liHerbknowKnowledgeEnhanced2020,liuNovelTransferLearning2022a,RoKEPG_BIBM_2023}, generative models have shown positive results in TCM prescription prediction. However, current models are beset with the following limitations:

Computational systems for prescription prediction in TCM model the complex relationships between symptoms and herbs \cite{yao2018topic}. Language generation techniques, treating prescription generation as a machine translation problem using sequence-to-sequence (seq2seq) models \cite{sutskeverSequenceSequenceLearning2014}, have shown promise. Despite positive results from generative models \cite{liuAttentiveherbNovelMethod2019,liHerbknowKnowledgeEnhanced2020,liuNovelTransferLearning2022a,RoKEPG_BIBM_2023}, current models still face the following limitations:
% % previous version -- dx 24-06-20 17:30
% Previous methods can be categorized into three types: (1): topic models, (2): graph models, and (3): language models. Language generation techniques, particularly promising in TCM prescription prediction, treat prescription generation as a machine translation problem and solve it using a sequence-to-sequence (seq2seq) model \cite{sutskeverSequenceSequenceLearning2014}. AttentiveHerb \cite{liuAttentiveherbNovelMethod2019} employs an attention mechanism that simulates clinical consultations by distinguishing between primary and secondary symptoms through symptom-focused attention. However, the scarcity of TCM datasets for prescriptions remains a significant challenge. To address this, Herb-Know \cite{liHerbknowKnowledgeEnhanced2020} introduced a seq2seq model enhanced with herbal knowledge in effect descriptions within a transformer framework. TCMBERT \cite{liuNovelTransferLearning2022a} further integrates a pre-trained BERT \cite{Devlin2019BERTPretraining} model with TCM documents to improve data extraction and modeling. Upon the pre-training on the TCM documents, RoKEPG \cite{RoKEPG_BIBM_2023} introduces extra knowledge of herbs and the corresponding attention mask matrix in the fine-tuning stage for prediction guidance. 
% \begin{itemize}
%     \item 
    \textbf{(1) Scarcity of clinical datasets.} 
    % Existing works often rely on datasets derived from classical documents rather than high-quality datasets from clinical records, which introduces noisy information. This results in models that cannot effectively provide personalized prescription recommendations \cite{zhang2021homogeneous}. Moreover, there is currently a lack of high-quality clinical data for prescription prediction \cite{dong2024presrecst}.
    Current models often rely on datasets derived from classical documents instead of high-quality clinical records, introducing noise and limiting the effectiveness of personalized prescription recommendations \cite{zhang2021homogeneous}. Moreover, there is a significant lack of high-quality clinical data for prescription prediction \cite{dong2024presrecst}.
    % \item 
    % \textbf{Sub-optimal Performance.} Existing models frequently yield results that do not meet expectations and may even mislead users due to their sub-optimal prediction accuracy \cite{dong2024presrecst}. However, relying solely on clinical data is insufficient to enhance model performance \cite{liu2023sequential}. It is necessary to integrate more advanced language model techniques to improve the representation and encoding capabilities of prescription prediction models.
   \textbf{(2) Sub-optimal Performance.} Existing models often underperform and may mislead users due to low prediction accuracy \cite{dong2024presrecst}. Relying solely on clinical data is insufficient to enhance model performance \cite{liu2023sequential}; integrating advanced language modeling techniques is necessary to improve the representation and encoding capabilities of prescription prediction models.
    % \item 
    \textbf{(3) Lack of herb dosage.} 
    % Existing models overlook the crucial prediction of dosage weights, a critical component in TCM for effective disease treatment \cite{yao2018topic,jinSyndromeawareHerbRecommendation2020a}. Additionally, there is a lack of evaluation metrics specifically for herb dosage prediction.
    Existing models overlook dosage weights, a critical component in TCM for effective treatment \cite{yao2018topic}. Furthermore, there is a lack of evaluation metrics specifically designed for herb dosage prediction.
% \end{itemize}

% Motivated by the robust predictive capabilities of large language models (LLMs), we develop a high-quality prescription dataset, \dataset, derived from clinical medical records, and propose \method—a novel LLM-based TCM prescription generation model (Fig. \ref{fig:work_flow}).
% leverage LLMs to construct an advanced prescription predictor. In this study, we introduce a high-quality prescription dataset \dataset derived from clinical medical records and propose \method (Fig. \ref{fig:work_flow}), a novel TCM prescription generation model based on LLMs. Our goal is to leverage the capabilities of LLMs to overcome the limitations of current methodologies. Specifically, we utilize a low-rank adaptation technique (LoRA) \cite{huLoRALowrankAdaptation2022} for efficient LLM fine-tuning. To take advantage of the order-agnostic nature of herb prescriptions, we implement data augmentation by randomizing the sequence of herbs in the training data. We validated our model on \dataset, achieving a Precision of 0.7951, a Recall of 0.8113, and an F1-score of 0.8031. This represents a significant improvement over the best performance achieved by previous methods. 
Motivated by the robust predictive capabilities of large language models (LLMs), we develop a high-quality prescription dataset, \dataset, derived from clinical medical records, and propose \method—a novel LLM-based TCM prescription generation model (Fig. \ref{fig:work_flow}). Specifically, we employ a low-rank adaptation technique (LoRA) \cite{huLoRALowrankAdaptation2022} for efficient fine-tuning of the LLM. Exploiting the order-agnostic nature of herb prescriptions, we implement data augmentation by randomizing the sequence of herbs in the training data. Validated on \dataset, our model achieves a precision of 0.7951, a recall of 0.8113, and an F1-score of 0.8031, significantly outperforming previous methods.
Our main contributions are outlined as follows:
\begin{figure*}
    \vspace{-3mm}
    \centering
    \includegraphics[width=0.9\linewidth]{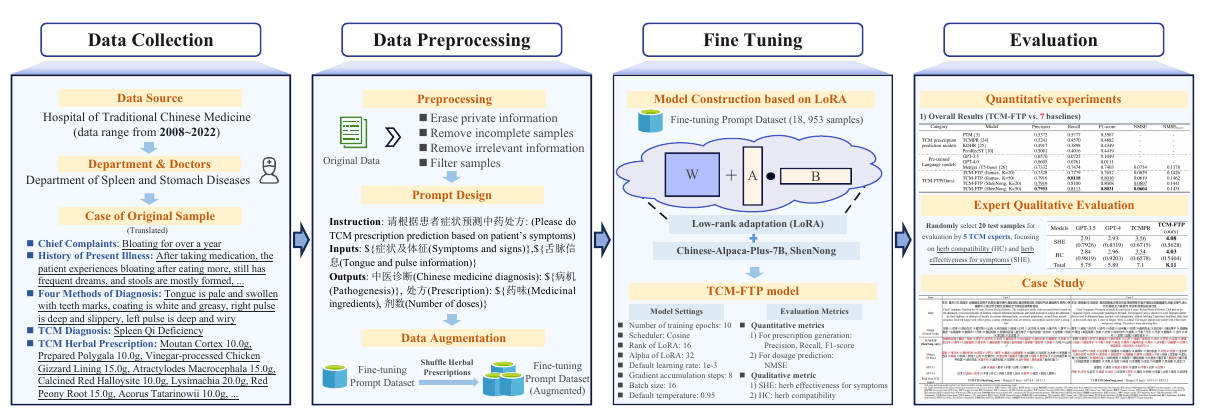}
    \caption{\textbf{Workflow of the TCM-FTP.} Our work consists of four parts: “Data Collection” involves gathering and organizing raw data; “Data Processing” includes data preprocessing, prompt design, and integrating data augmentation to create a fine-tuning dataset; “Fine-tuning” utilizes the ShenNong LLM and LoRA technique to optimize the model; and “Evaluation” assesses the outcomes using both quantitative and qualitative evaluation metrics.}
    \label{fig:work_flow}
    \vspace{-3mm}
\end{figure*}

% original text for figure1 -- dx 24-06-20 18:19
% The TCM clinical prescription data items are specifically collected with a focus on digestive system disorders. The preprocessing of these collected items involves several steps: erasing private information, removing incomplete samples, excluding irrelevant information, and filtering out samples that pertain only to first-visit patient records. The processed samples are then compiled into an instruction dataset and fed into LLMs for fine-tuning. A low-rank adaptation technique \cite{huLoRALowrankAdaptation2022} is employed for parameter-efficient fine-tuning, where the original parameters are kept frozen and the learnable parameters are represented by low-rank matrices.

\begin{itemize}[leftmargin=0pt]
% \item We construct a TCM clinical diagnostic dataset \dataset derived from practical records in real-world TCM clinical scenarios. This dataset serves as a foundational resource for understanding the nuanced relationships between symptoms and effective prescriptions in TCM.
\item We construct \dataset, a TCM clinical diagnostic dataset derived from real-world clinical records, serving as a foundational resource for understanding the nuanced relationships between symptoms and effective prescriptions in TCM.
\item 
% We introduce \method, a novel LLM-driven method for generating TCM prescriptions. Our model excels at capturing the intricate interplay between symptoms and herbal prescriptions, marking a significant advancement in TCM prescription prediction. Additionally, we have designed the NMSE evaluation metric for dosage prediction. To the best of our knowledge, our work is the first in the prescription prediction field to design dosage prediction metrics tailored for real-world scenarios.
We introduce \method, a novel LLM-driven method for generating TCM prescriptions that excels at capturing the intricate interplay between symptoms and herbal treatments, marking a significant advancement in prescription prediction. Additionally, we introduce the Normalized Mean Square Error (NMSE) as an evaluation metric for dosage prediction, pioneering the first dosage prediction metric tailored to real-world scenarios in this field.

% \item We empirically validate \method against existing models on our dataset, showcasing its superior performance in predicting herbs and dosages. Additionally, we conduct a thorough analysis of factors such as the impact of foundational models, learning rates, the number of herb permutations, and the decoding parameters in inference. Expert case studies by TCM professionals further provide valuable insights into the adaptability and robustness of our innovative approach.

\item We empirically validate \method against existing models on our dataset, demonstrating its superior performance in predicting herbs and dosages. Expert case studies by TCM professionals further provide valuable insights into the adaptability and robustness of our innovative approach.

\end{itemize}

\section{BACKGROUND AND SIGNIFICANCE}

\subsection{Problem Definition}
We now formally define the prescription prediction problem.  
 Given a prescription dataset $\mathcal{P}_{\text{train}}$, an element in $\mathcal{P}_{\text{train}}$ consists of a symptom description $s$ and its corresponding prescription $\{h_i,w_i\}^{i\in[k]}$. Here, $h$ denotes the herb name, $w$ represents the herb dosage, and $[k]$ signifies a list ranging from 1 to $k$. Our objective is to train a model $\mathcal{M}$ such that $\mathcal{M}(s)$ reproduce $\{h_i,w_i\}^{i\in[k]}$ accurately. For alignment with the language generation task, we concatenate $\{h_i,w_i\}^{i\in[k]}$ using a comma separator to form a single sentence. Concrete examples are presented in Fig. \ref{fig:work_flow}.

\subsection{TCM Herbal Prescription Prediction}
Research on TCM prescription prediction mainly falls into three categories: topic model-based, graph model-based, and language model-based approaches. Topic model-based approaches treat relationships between symptoms and herbs as that of documents and topics \cite{wangKnowledgeGraphEnhanced2019}. These approaches rely on statistical relationships, lacking a semantic understanding of symptoms. Graph-based approaches construct a medical knowledge graph to model the relationships between symptoms and herbs \cite{yangMultilayerInformationFusion2022}. However, these approaches also lack consecutive semantic information regarding symptoms. 
% Language model-based approaches are a more promising way to model the complicated relationships between symptoms and herbs.
Language model-based prescription prediction models take patient symptom descriptions as input and generate herbal prescriptions sequentially. TCM Translator \cite{wangTcmTranslatorSequence2019} uses transformer architectures to distill context vectors from symptoms and LSTM
% \cite{Hochreiter1997LongShortTerm} 
as the decoder. AttentiveHerb \cite{liuAttentiveherbNovelMethod2019} employs a seq2seq
% \cite{sutskeverSequenceSequenceLearning2014}  
model with dual attention mechanisms to distinguish primary from secondary symptoms and map herb-symptom interactions using clinical data. Herb-Know \cite{liHerbknowKnowledgeEnhanced2020} utilizes herb descriptions to model associations with symptoms and evaluates whether herb effects align with symptom descriptions, and other related work like TCMBERT \cite{liuNovelTransferLearning2022a}, RoKEPG \cite{RoKEPG_BIBM_2023} also have shown their effectiveness on TCM prescription prediction.

\subsection{Large Language Models in TCM}

LLMs have advanced significantly in NLP tasks, with models like ChatGPT and GPT-4 \cite{achiam2023gpt} drawing attention, though OpenAI has not disclosed architectural details. Open-source LLMs, such as LLaMA \cite{touvronLLaMAOpenEfficient2023} and ChatGLM \cite{Zeng2023GLM130BOpen}, have spurred research interest, though they often struggle in specialized fields like medicine or finance. Supervised fine-tuning has become the standard to enhance these models with domain-specific knowledge. In TCM, tailored models have emerged. Bentsao \cite{wangHuaTuoTuningLLaMA2023a} fine-tunes LLaMA using structured and unstructured knowledge from CMeKG \cite{byambasurenPreliminaryStudyConstruction2019}. Huatuo \cite{zhangHuatuoGPTTamingLanguage2023a} incorporates ChatGPT data with Reinforcement Learning from Artificial Intelligence Feedback (RLAIF) and Proximal Policy Optimization (PPO). Zhongjing \cite{yangZhongjingEnhancingChinese2023a} applies continual pretraining with supervised fine-tuning on a Chinese medical dialogue dataset, integrating Reinforcement Learning from Human Feedback (RLHF). ShenNong \cite{zhu2023ChatMedShenNong} combines LoRA-based fine-tuning on instructional data and TCM knowledge, supported by over 110,000 instructions.

 \subsection{Parameter Efficient Fine-Tuning}
 
 Parameter Efficient Fine-Tuning (PEFT) utilizes a small amount of parameters to fine-tune a large language model effectively. Assuming there is a pretrained model $f_{\theta}(y|x)$, PEFT seeks to adjust a limited number of parameters, $\Delta \theta$, such that $|\theta| \gg |\Delta\theta|$. In contrast to conventional fine-tuning, which updates all parameters, denoted as $|\theta|=|\Delta\theta|$, and requires significant computational resources, PEFT selectively updates a limited number of learnable parameters to achieve results comparable to complete fine-tuning.  In this work, we employ LoRA \cite{huLoRALowrankAdaptation2022} to efficiently fine-tune LLMs on the target dataset.

\section{MATERIALS AND METHODS}
In this study, as illustrated in Fig. \ref{fig:work_flow}, we gather a high-quality prescription dataset \dataset and propose a PEFT approach for prescription prediction \method. \dataset is comprised of practical medical records collected from specialists in digestive system disorders. \method utilizes supervised fine-tuning to train the model on \dataset, incorporating LoRA technique and an effective data augmentation technique, which involves permuting the herbs in the prescriptions.

% The dataset is built by the practices from  TCM experts. We employ low random adaptation (LoRA) technique in tuning with less computational resource demands. Further, we explore ingredient shuffling as an effective data augmentation. The following parts are arranged: we first introduce a prescription dataset built by collecting practical first-visit patient records, followed by the problem definition. Next, we elaborate on the fine-tuning process for \method, emphasizing underfitting and an order-agnostic property in the generated prescriptions. We also list baselines and provide the details of fine-tuning experiment setup for the purpose of reproducing.  The entire workflow can be found in Figure \ref{fig:work_flow}.

\subsection{Datasets}
\label{sec:dataset}
\subsubsection{Data collection} We collect outpatient medical record data from practical patient records generated by specialists in TCM hospital \footnote{by Dr. Runshun Zhang in Guang’anmen Hospital, China} over a span from 2008 to 2022. The prescriptions specifically focus on digestion system disorders.

\subsubsection{Data processing} Initially, we remove the incomplete data items and erase any privacy information. Subsequently, we exclude irrelevant information, retaining only essential information for prescription prediction. Specifically, we keep the chief complaint, medical history, and tongue-coating details from the symptom descriptions, as well as the names and dosages of herbs in the prescriptions.

\subsubsection{Data statistics}
\begin{figure}[htbp]
    \VspaceBefore
  \centering
  \begin{subfigure}{0.4 \linewidth}
    \centering
    \includegraphics[width=\linewidth]{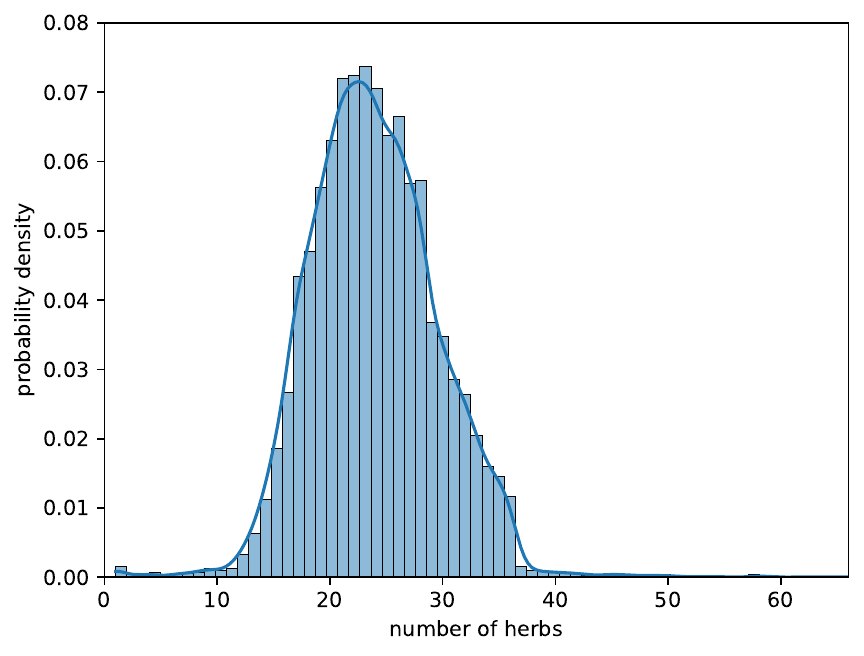}
    \caption{Training dataset}
  \end{subfigure}
  \hfill
  \begin{subfigure}{0.4 \linewidth}
    \centering
    \includegraphics[width=\linewidth]{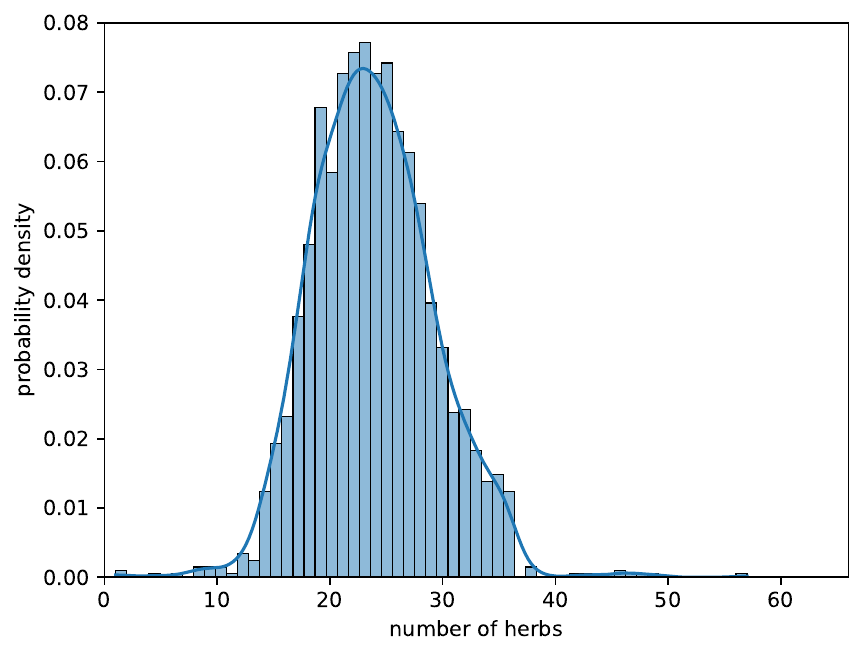}
    \caption{Test dataset}
  \end{subfigure}
  \caption{\textbf{Distribution of the number of herbs in prescriptions} The blue curves represent kernel density estimates.}
  \label{fig:distribution_num_herbs}
  % \VspaceAfter
  \vspace{-6mm}
\end{figure}

We present the processed data statistics in Table \ref{tab:dataset_statistics}. Prescription items are randomly divided into training (90\%) and testing (10\%) datasets. The training dataset comprises 16,896 samples with an average of 23.92 herbs per prescription, while the test dataset includes 2,057 samples averaging 23.86 herbs per prescription. Fig. \ref{fig:distribution_num_herbs} illustrates the distributions of herb counts in the training and testing datasets.

% We show the statistics of processed data items in Table \ref{tab:dataset_statistics}. We randomly separate prescription items into train (90\%) and test (10\%) datasets. The training dataset contains 16,896 samples with an average number of herbs in a prescription of 23.92, while the test dataset contains 2,057 samples with an average number of herbs in a prescription of 23.86. In Figure \ref{fig:distribution_num_herbs}, we present the distributions of the number of herbs in train and test datasets.  

% Table generated by Excel2LaTeX from sheet 'Sheet 1'
\begin{table}[htbp]
    \VspaceBefore
  \centering
  \small
  \caption{\textbf{Statistics of training and test datasets} The table displays the median, mean, and standard deviation for the number of herbs per prescription. Category: Number of distinct herbs in the dataset.}
    \label{tab:dataset_statistics}%
    \small
    \renewcommand{\arraystretch}{0.8}
    \begin{tabularx}{\linewidth}{l*{5}{>{\centering\arraybackslash}X}}
    
		\specialrule{1pt}{0pt}{0pt}
    dataset & size & category & median & mean & std \\
    \midrule
    training & 16,896 & 674  & 24  & 23.92 & 5.69 \\
    test & 2,057 & 533  & 24   & 23.86 & 5.44 \\
		\specialrule{1pt}{0pt}{0pt}
    \end{tabularx}%
  \VspaceAfter
  \vspace{-2.5mm}
\end{table}%
% revised statistics with new investigation. 

% \clearpage

\subsection{\method}
\label{sec:method}
% \method v1.1

To better model the intricate relationships between symptoms and herbs, we propose \method, which employs a pre-trained LLM coupled with an auto-regressive mechanism for TCM prescription prediction. For efficient fine-tuning of the LLM, we employ the LoRA technique\cite{huLoRALowrankAdaptation2022}, which optimizes the LLM with limited parameters and computational resources. To enhance the dataset, we permute the herbs in prescriptions to leverage the order-agnostic property inherent in TCM prescriptions. 

Given a prescription dataset $\mathcal{P}_{train}$ with $\{s,\{h_i,w_i\}^{i\in [k]}\}$, our primary goal is to use a language model to minimize the negative logarithm of the estimation of conditional probability: 
\begin{equation}
   \min\limits_{\theta} \mathop{\mathbb{E}}\limits_{s,\{h_i,w_i\}^{i\in [k]} \sim \mathcal{P}_{train}} \sum\limits_{i}^k -\log P((h_i,w_i)|h_{<i},w_{<i};s; \theta),
\end{equation}
where $s$ is the symptom description, $k$ is the number of herbs in the prescription, and $(h_i, w_i)$ represents the i-th (herb, dosage) pair with $i\in\{1,2,...,k\}$. $h_{<i}$ and $w_{<i}$ represent a set of herbs and a set of weights with index less than $i$, respectively. $\theta$ represents our model parameters. 

We concatenate the herbs and dosages into one string to use the auto-regressive mechanism in LLMs and get the goal:
% \begin{equation}
%  \max\limits_{\theta}   \mathop{\mathbb{E}}\limits_{s,\{h_i,w_i\}^{i\in [k]} \sim \mathcal{P}_{train}} \log P(\textsc{cat}(\{h_i,w_i\}^{i\in [k]})|s; \theta),
% \end{equation}
 % Our auto-regressive target is then: 
\begin{equation}
 \min\limits_{\theta}   \mathop{\mathbb{E}}\limits_{\substack{s,\{h_i,w_i\}^{i\in [k]} \sim \mathcal{P}_{train}\\ y=\textsc{cat}(\{h_i,w_i\}^{i\in [k]})}} \sum\limits_{j=1}^{|y|}  -\log P(y_j|\{y_{<j}\};s; \theta),
\end{equation}
where $\{y_{<j}\}$ denotes the set of the tokens from index $1$ to index $j-1$ in $y$, and $|y|$ represents the number of tokens in $y$. $\textsc{cat}(\{h_i,w_i\}^{i\in [k]})$ is the concatenated string. 

% {\color{red} TODO: further write in auto-regressive form. 
% }

% s,\{h_i,w_i\}^{i\in [k]} \sim \mathcal{P}_{train}
% Our objective function is: 
% \[
% \mathcal{L} = - \sum\limits_{x,y \in {\mathcal{P}_{train}}} {\log p\left( {y|x;{\theta _0} + \Delta \theta } \right)}, 
% \]
% where $p(y|x;{\theta _0})$ signifies the likelihood produced by the initial pre-trained model parameterized by $\theta_0$. $\Delta \theta$ denotes the adjustable parameters.

% {\color{red} TODO: elaborate the model}

\subsubsection{Low-rank Adaptation}
\label{sec:low-rank Adaptation}

We employ the LoRA \cite{huLoRALowrankAdaptation2022} to reduce computational resources. Towards $f_\theta$, we encounter two primary types of parametric functions within our model framework: linear and embedding functions, defined as: 
 \begin{align}
 {\mathcal{F}_{\text{linear}}}(x) = {W_l} \cdot x, 
 	{\mathcal{F}_{\text{emb}}}(x) = \text{EMB}\left( x;{W_e} \right),
 \end{align}
 where \text{EMB}$(x;\cdot)$ represents the embedding operator that selects the $x$-th column from the specified matrix, and $W_l$ and $W_e$ denote the parameters for the linear and embedding functions, respectively. For fine-tuning, we introduce updates via low-rank matrices $A$ and $B$ to modify the original functions:
\begin{equation}
\begin{split}
    {\mathcal{F}_{\text{linear}}}{(x)^\prime } &= {\mathcal{F}_{\text{linear}}}(x) + {A_l} \cdot {B_l} \cdot x, \\
 	{\mathcal{F}_{\text{emb}}}{(x)^\prime } &= {\mathcal{F}_{\text{emb}}}(x) + {A_e} \cdot \text{EMB}\left( x;{B_e} \right),
\end{split}
\end{equation}
 where $A_l$, $B_l$, $A_e$, and $B_e$ are the learnable parameters. With the updates to the low-rank matrices represented by $\Delta\theta$, our goal is:
\begin{equation}
 \min\limits_{\Delta\theta}   \mathop{\mathbb{E}}\limits_{\substack{s,\{h_i,w_i\}^{i\in [k]} \sim \mathcal{P}_{train}\\ y=\textsc{cat}(\{h_i,w_i\}^{i\in [k]}}} \sum\limits_{j=1}^{|y|}  -\log P(y_j|\{y_{<j}\};s; \theta +\Delta\theta).
\end{equation}

% refrains from adding new modules or altering architectures, instead learning low-rank matrices that initiate model-wide updates, based on the premise that representations occupy low-rank spaces.
% \begin{equation}
%     \begin{aligned}
%  	{\mathcal{F}_{\text{linear}}}{(x)^\prime } &= {\mathcal{F}_{\text{linear}}}(x) + {A_l} \cdot {B_l} \cdot x, \\
%  	{\mathcal{F}_{\text{emb}}}{(x)^\prime } &= {\mathcal{F}_{\text{emb}}}(x) + {A_e} \cdot \text{EMB}\left( x;{B_e} \right),
%  \end{aligned}
% \end{equation}

% \begin{equation}
%  \max\limits_{\Delta\theta}   \mathop{\mathbb{E}}\limits_{s,\{h_i,w_i\}^{i\in [k]} \sim \mathcal{P}_{train}} P(\textsc{cat}_i(h_i,w_i)|s; \theta+\Delta\theta),
% \end{equation}

\subsubsection{Order-Agnostic Property}
\label{sec:order-agnostic property}
Recognizing the order-agnostic characteristic of herbs in TCM prescriptions, we implement data augmentation by permuting the herbs in the prescriptions. Given a prescription sample $\{s, \{h_i, w_i\}^{i \in [k]} \}$, we define the permuted herb sequence as \( \textsc{cat} \mathop{\shuffle}(\{h_i, w_i\}^{i \in [k]}) \). The sequence resulting from the permutation, \( \mathop{\shuffle}(\{h_i, w_i\}^{i \in [k]}) \), is represented as \( \{h_{r_i}, w_{r_i}\}^{i \in [k]} \), where \( r_i \) denotes the indices after shuffling. After the herb permutation, our final goal becomes:
\begin{equation}
 \min\limits_{\Delta\theta} \sum\limits_{t=1} ^K  \mathop{\mathbb{E}}\limits_{\substack{s,\{h_i,w_i\}^{i\in [k]} \sim \mathcal{P}_{train}\\ y=\textsc{cat} \mathop{\shuffle}(\{h_i,w_i\}^{i\in [k]})}} \sum\limits_{j=1}^{|y|}  -\log P(y_j|\{y_{<j}\};s; \theta +\Delta\theta),
\end{equation}
where $K$ is the number of permutation times. $t$ refers to different permutation index since permutation function $\shuffle$ is nondeterministic at each run.  

% \TODO{Check consistency of math symbols.}
% \begin{equation}
%  \max\limits_{\Delta\theta}   \mathop{\mathbb{E}}\limits_{s,\{h_i,w_i\}^{i\in [k]} \sim \mathcal{P}_{train}} P(\textsc{cat}_i\mathop{\shuffle}\limits_{i}^k{(h_i,w_i)}|s; \theta+\Delta\theta),
% \end{equation}

% \subsubsection{Data distribution divergence and underfitting}

% Unlike conventional NLP tasks prioritizing answers that align with human preferences, TCM prescription prediction is indifferent to herb order. This indifference leads to substantial divergence in target data distributions. Consequently, there's a pronounced disparity between the pre-trained and target models, necessitating extended training.

% In response, we increased the learning rate and number of training epochs to counteract distribution divergence and underfitting. Comprehensive empirical studies were conducted to evaluate the effects of learning rate and the number of herb permutations, referring to Sec. \ref{sec:parameter_analysis}.

% Acknowledging the order-agnostic nature of herbs in TCM prescriptions, we implemented data augmentation by randomizing the sequence of herbs. Given a prescription pair $s, \{h_i,g_i\}^{i\in[k]}$, the pair $\{h_i,g_i\}^{i\in[k]}$ is randomized to $\{h_{s_i},g_{s_i}\}^{i \in [k]}$, where $s_1,s_2,...,s_k$ are randomly permuted indices from $1,2,...,k$. This modified pair $s, \{h_{s_i},g_{s_i}\}^{i\in[k]}$ is added to the training set for each original data point. We replicate this augmentation process $m$ times and define the random permutation operation as $\shuffle$. Our final objective becomes: 

\subsection{Baselines}
To demonstrate the performance of our model, we compare it with several baselines, including TCM prescription prediction models (PTM \cite{yao2018topic}, TCMPR \cite{Dong2021TCMPRTCMPrescription}, KDHR \cite{yang2022multi}, and PresRecST \cite{dong2024presrecst}), and pre-trained language models (Mengzi (T5-base) \cite{zhang2021mengzi}, GPT-3.5 and GPT-4.0 \cite{achiam2023gpt}).

\subsection{Experimental setup}
% {\color{red} TODO: secify the hyper parameters}

We implement the fine-tuning process with the Transformers package from Hugging Face\footnote{\url{https://huggingface.co/docs/transformers}}. The number of training epochs is set to 10. We adopt a cosine scheduler, with the LoRA rank set to 16 and the LoRA alpha to 32.  We consider the number of herb permutations, K, as 20 and 50, respectively. The default learning rate is set to $1\times10^{-3}$. We leverage the gradient accumulation technique to increase the batch size, with 8 accumulation steps and a batch size of 16. We employ two foundation models Chinese-Alpaca-Plus-7B \footnote{\url{https://github.com/ymcui/Chinese-LLaMA-Alpaca/tree/main}} and ShenNong \footnote{\url{https://huggingface.co/michaelwzhu/ShenNong-TCM-LLM}}.  Chinese-Alpaca-Plus-7B is a variant of LLaMA \cite{touvronLLaMAOpenEfficient2023} with continual pre-training and supervised fine-tuning on Chinese corpus, denoted as LLaMA+ for simplicity. ShenNong is a further refinement of LLaMA+, fine-tuned with TCM instruction datasets. We use an 8-V100 GPU machine for fine-tuning and the running time for \method ($K=50$) is 146 hours. During the inference stage, We employ top-k and top-p combinations for decoding, setting the top-k to 50 and the top-p to 0.7, with a default temperature of 0.95.

\subsection{Evaluation Metrics}
% For the proposed TCM-FTP, we evaluate it from both quantitative and qualitative perspectives, including the following evaluation metrics.

\textbf{Quantitative Evaluation.}
% We use precision, recall, and F1-score as herb prediction metrics and designed NMSE for herb dosage evaluation. 

\begin{itemize}[leftmargin=0pt]
    \item Precision, recall, and F1-score: Precision is the proportion of correct herb predictions, recall is the proportion within true herbs, and F1-score is their harmonic mean.
	% \item Precision: The proportion of correctly predicted herbs out of all predicted herbs, reflecting the accuracy of positive predictions.
	% \item Recall: The proportion of correctly predicted herbs out of all herbs in the ground truth, indicating the model's ability to identify all relevant items.
	% \item F1-score: The harmonic mean of precision and recall, giving a balanced measure of overall accuracy.
	\item Normalized mean square error (NMSE): We design a metric to evaluate the accuracy of predicted dosage by normalizing the squared differences using the original weights. For a given prescription $p$ and a predicted prescription $\hat{p}$, we suppose they are composed of a set of pairs of a herb and a dosage, $p=\{(h,w)\}$, where $h,w$ refer to herb and dosage respectively, and $\hat{p}=\{(h^{\prime},w^{\prime})\}$ indicates the generated result from model. The NMSE is defined as:
  $NMSE = \frac{1}{Z}\sum_{\substack{(h, w) \in p, \\\left(h^{\prime}, w^{\prime}\right) \in \hat{p}}} \mathbf{1}\left[h=h^{\prime}\right]\left(\frac{w^{\prime}-w}{w}\right)^2,$
where $Z$ is the number of correctly predicted herbs, and $\mathbf{1}[\cdot]$ is the indicator function. For baseline, we use the average dosage of each herb from the training data for dosage predictions. 
% For herbs unseen in the training data, the dosage is predicted as the average dosage of all known herbs. 
% This approach is referred to as NMSE$_{base}$ in the NMSE calculations.

\end{itemize}

\textbf{Qualitative Evaluation.} Existing quantitative evaluation metrics assess model quality based solely on sample labels, neglecting the compatibility and symptom-specific effectiveness of prescriptions. To comprehensively evaluate our model, we engaged five TCM experts to conduct an expert qualitative evaluation (EQE) of selected prescriptions generated by our model. Each doctor independently assessed the prescriptions for herbal effectiveness in treating symptoms (SHE) and herbal compatibility (HC), assigning scores on a scale of 0 to 5. Higher scores indicate greater effectiveness or compatibility.
% Existing quantitative evaluation metrics only assess model quality based on sample labels, without considering the compatibility and symptom-specific effectiveness of the prescriptions. Therefore, these metrics cannot fully reflect the model's capabilities. To evaluate our model more comprehensively, we invited five TCM doctors from TCM Hospital to independently perform a expert qualitative evaluation (EQE) of some prescriptions generated by our model. Each doctor evaluated the prescriptions in terms of herbal effectiveness for treating symptoms (SHE) and herbal compatibility (HC), with scores ranging from 0 to 5. Higher scores indicate better effectiveness or compatibility.

\section{RESULTS AND DISCUSSION}
%To empirically verify the effectiveness of our model, we design experiments to answer following questions:
%\begin{itemize}
%	\item How does our approach compare in performance to existing methods?
%	\item Our model can predict weights of herbs in prescriptions. How accurate is the weight prediction?
%	\item How do different hyperparameters, encompassing both training and inference parameters, influence the results?
%	\item How is our method adept at extracting classic formulae?
%\end{itemize}

\subsection{Overall Results}

\begin{table*}[t]
    \vspace{-3mm}
	\centering
 \small
	\caption{\textbf{Prediction results of herbs and dosages in prescriptions} \method is evaluated with different foundation models.}
 \vspace{-2mm}
  \label{tab:main_result}%
      \small
      
\resizebox{1\linewidth}{!}{

    \begin{tabularx}{\textwidth}{ll*{5}{>{\centering\arraybackslash}X}}
		\specialrule{1pt}{0pt}{0pt}
        \makecell[c]{Category}&\makecell[c]{Model} & Precision & Recall& F1-score& NMSE & NMSE$_{base}$\\
        \midrule
       %  \multirow{3}{*}{\makecell[c]{MLC models \& \\ topic models}}& MLKNN \cite{Zhang2007MLKNNlazy} & 0.5365 & 0.4626 & 0.4968 & - & - \\
       %  &LinkLDA \cite{Jiang2012Usinglinktopic} & 0.5267 & 0.4572 & 0.4895 & - & - \\
       % & LinkPLSALDA \cite{Nallapati2008LinkPLSALDA} & 0.5311 & 0.4614 & 0.4938 & - & - \\
       %  \hdashline
        \multirow{4}{*}{\makecell[c]{TCM prescription \\ prediction models}} & PTM \cite{yao2018topic} & 0.5372 & 0.5777 & 0.5567 & - & - \\
        &TCMPR \cite{Dong2021TCMPRTCMPrescription} & 0.5241 & 0.4570 & 0.4882 & - & - \\
        &KDHR \cite{yang2022multi} & 0.4917 & 0.3898 & 0.4349 & - & - \\
        &PresRecST \cite{dong2024presrecst} & 0.5061 & 0.4016 & 0.4419 & - & - \\
        \hdashline
       \multirow{3}{*}{\makecell[c]{Pre-trained \\ Language models}} & GPT-3.5 & 0.0570 & 0.0725 & 0.1049 & - & - \\
       & GPT-4.0 & 0.0605 & 0.0761 & 0.0111 & - & - \\
       & Mengzi (T5-base) \cite{zhang2021mengzi} &0.7332 &  0.7474 & 0.7403 & 0.0754 & 0.1378\\
        \hdashline
        \multirow{4}{*}{\makecell[c]{TCM-FTP(Ours)}} &\method(llama+, K=20)  & 0.7528 & 0.7779 & 0.7652 & 0.0829 & 0.1426\\
        &\method(llama+, K=50) & 0.7916 & \textbf{0.8118} & \underline{0.8016} & 0.0619 & 0.1462 \\
        &\method(ShenNong, K=20) &\underline{0.7919} & 0.8100 & 0.8008 & \underline{0.0607} & 0.1441\\
        &\method(ShenNong, K=50) &\textbf{0.7951} & \underline{0.8113} & \textbf{0.8031} & \textbf{0.0604} & 0.1431 \\
		\specialrule{1pt}{0pt}{0pt}
    \end{tabularx}
    }
    \vspace{-6mm}
\end{table*}%

As shown in Table \ref{tab:main_result}, the proposed \method outperforms all baseline models in the herb prediction task on \dataset, achieving an F1-score of 0.8016 using LLaMA+ as the foundation model and 0.8031 using ShenNong. This highlights the superior capability of \method in modeling the intricate relationships between symptoms and herbs in prescriptions. Unlike previous approaches, \method also includes herb dosage prediction, which is crucial in TCM due to the significant impact of dosage combinations. As shown in Table \ref{tab:main_result}, \method achieves a much lower NMSE compared to the baseline using average statistics (0.0604 for \method versus 0.0754 for Mengzi (T5-base) \cite{zhang2021mengzi}). This enhances the practicality of \method.

% Our proposed \method shows significant advantages over various baseline models across all aspects. 
Compared to the pre-trained language model Mengzi (T5-base) \cite{zhang2021mengzi}, \method significantly enhances prediction performance with improved precision (from 0.7332 to 0.7951), recall (from 0.7474 to 0.8113), and F1-score (from 0.7403 to 0.8031). Performances on GPT-3.5 and GPT-4.0 \cite{achiam2023gpt} were notably poor, indicating the limitations of general LLMs in tasks requiring specialized knowledge. In comparison to other TCM prescription prediction models, \method outperforms PTM \cite{yao2018topic}, which leads among these baselines, followed by TCMPR \cite{Dong2021TCMPRTCMPrescription} in accuracy, with KDHR \cite{yang2022multi} and PresRecST \cite{dong2024presrecst} showing poorer performance. PTM, a topic model-based approach, is computationally complex, while TCMPR, KDHR, and PresRecST are graph-based models sensitive to graph-related factors. \method excels these approaches in herb recommendations due to advanced language modeling capability.

\subsection{Expert Qualitative Evaluation}
Since conducting expert qualitative evaluation (EQE) is time-consuming and labor-intensive for doctors, this study randomly selected 20 data points from the test set for evaluation. As shown in Table \ref{tab:eqe_result}, the TCM-FTP model outperformed other baseline models across all metrics, achieving an average SHE score of 4.08 (standard deviation 0.5628) and an average HC score of 4.03 (standard deviation 0.5404). This indicates that TCM-FTP can generate effective prescriptions that adhere to TCM compatibility principles. In contrast, GPT-3.5 and GPT-4.0 \cite{achiam2023gpt} scored below 3 for both SHE and HC, with total scores of 5.75 and 5.89, respectively, significantly lower than TCM-FTP and TCMPR \cite{Dong2021TCMPRTCMPrescription}. This demonstrates that GPT-3.5 and GPT-4.0 perform poorly in generating prescriptions that adhere to TCM principles. TCMPR performed relatively well, with SHE and HC scores of 3.56 (standard deviation 0.6715) and 3.54 (standard deviation 0.6578), respectively, but still fell short of TCM-FTP. Overall, TCM-FTP excelled in prescription generation, receiving higher approval from doctors and showing less score variability, indicating better stability and robustness.

\begin{table}[h]
    \VspaceBefore
    \caption{\textbf{Comparative Results of expert evaluation.} The table records the mean scores (standard deviation).}
    \centering
    \footnotesize
    \begin{tabular}{c c c c c }
    \specialrule{1pt}{0pt}{0pt}
        Models & GPT-3.5 & GPT-4.0 \cite{achiam2023gpt} & TCMPR \cite{Dong2021TCMPRTCMPrescription} & \textbf{TCM-FTP}(ours) \\ \specialrule{0.6pt}{0pt}{0pt}
        SHE & \makecell[c]{2.91 \\(0.7926)} & \makecell[c]{2.93 \\(0.8319)} & \makecell[c]{\underline{3.56} \\(0.6715)} & \makecell[c]{\textbf{4.08} \\(0.5628)} \\ 
        HC & \makecell[c]{2.84 \\(0.9819)} & \makecell[c]{2.96 \\(0.9203)} & \makecell[c]{\underline{3.54} \\(0.6578)} & \makecell[c]{\textbf{4.03} \\(0.5404)} \\ 
        Total & 5.75 & 5.89 & 7.1 & \textbf{8.11} \\ \specialrule{1pt}{0pt}{0pt}
    \end{tabular}
    \label{tab:eqe_result}
    \vspace{-3mm}
\end{table}

\begin{figure*}[!htbp]
    \vspace{-3mm}
    \centering
        \caption{\textbf{Two Specific Case Analyses.} We present two specific test cases, including inputs, outputs, and predictions of the models with evaluations of experts. Herb names/dosage weights are marked in red for corrected ones.}
    \label{fig:case}
    \includegraphics[width=0.9\linewidth]{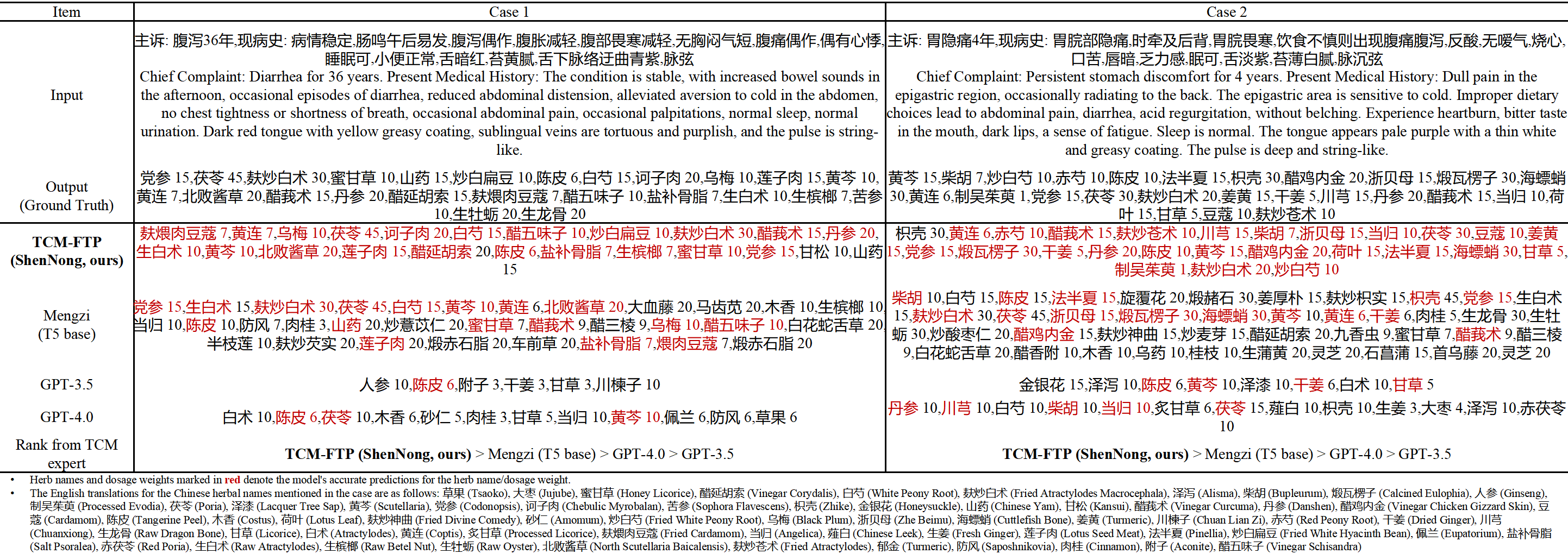}
    \vspace{-5mm}
\end{figure*}
\subsection{Case Study}
In order to visually showcase the predictive performance and capabilities of our \method, we obtain the predicted herb results formed by each model on the test set and had them evaluated by TCM experts. Fig. \ref{fig:case} illustrates the results for two cases, including the input (chief complaints, present medical history), output (actual prescriptions provided by the doctor), and the predicted prescription results from \method (ShenNong) and Mengzi (T5 base) \cite{zhang2021mengzi} models, as well as GPT-3.5 and GPT-4.0 \cite{achiam2023gpt} predictions. Herb names/dosage weights marked in red indicate that the model's prediction match the actual prescription provided by the doctor. Additionally, professional evaluations of each model are presented, ranking them using the “\textgreater” symbol (with models ranked higher indicating better performance). Results show that \method (ShenNong) achieves strong prediction accuracy, closely matching doctors' herb labels and dosages after fine-tuning, with clinically relevant dosage alignment. In contrast, GPT-3.5 and GPT-4.0 showed poor predictions due to a lack of specialized TCM training. This underscores the value of fine-tuning LLMs on high-quality, domain-specific data for optimal performance in specialized fields.

\section{CONCLUSION}
% v1.0
To deal with the lack of high-quality datasets and to improve the performance in TCM prescription predictions, we build a TCM prescription dataset \dataset and propose \method to fine-tune LLMs to predict herbs with the corresponding dosages. 
% We construct \dataset from practical clinical records by focusing on digestion disorder diseases. \method employs a low-rank adaptation for computational and storage efficiency and adapts a data augmentation by randomly permutating the order of herbs in prescriptions. 
The experimental results reveal the remarkable effectiveness of \method, surpassing previous methods by large margins in precision, recall, and F1-score. Additionally, our method achieved the best results in NMSE, effectively forming accurate herb and dosage predictions. In future work, we will continue to incorporate domain knowledge into model construction to further enhance performance, aiming to develop a practically usable prescription prediction model.

\section{ACKNOWLEDGEMENT}
This work is partially supported by the Natural Science Foundation of Beijing (No. L232033), the National Natural Science Foundation of China (Nos. 82374302 and U23B2062), the Fundamental Research Funds for the Central Universities (No. 2024YJS032) and Hong Kong Research Grants Council under grant (No. 16204920). We gratefully acknowledge the support of NVIDIA Corporation with the GPU computational resources used for this research.

\bibliographystyle{IEEEtran}
\bibliography{manuscript}

\clearpage

\end{document}